\documentclass{article}

\usepackage[preprint, nonatbib]{neurips_2024}
\usepackage{amsmath}

\usepackage[utf8]{inputenc} % allow utf-8 input
\usepackage[T1]{fontenc}    % use 8-bit T1 fonts
\usepackage{hyperref}       % hyperlinks
\usepackage{url}            % simple URL typesetting
\usepackage{booktabs}       % professional-quality tables
\usepackage{amsfonts}       % blackboard math symbols
\usepackage{nicefrac}       % compact symbols for 1/2, etc.
\usepackage{microtype}      % microtypography
\usepackage{xcolor}         % colors
\usepackage{graphicx}
\usepackage{multirow}
\usepackage{makecell}

\title{Self-supervised Video Instance Segmentation Can Boost Geographic Entity Alignment in Historical Maps}

\author{%
  Xue Xia \\
  ETH Zurich \\
  \And
  Randall Balestriero \\
  Brown University \\
  \AND
  Tao Zhang \\
  Wuhan University \\
  \And
  Lorenz Hurni \\
  ETH Zurich \\
}

\begin{document}

\maketitle

\begin{abstract}
  Tracking geographic entities from historical maps, such as buildings, offers valuable insights into cultural heritage, urbanization patterns, environmental changes, and various historical research endeavors. However, linking these entities across diverse maps remains a persistent challenge for researchers. Traditionally, this has been addressed through a two-step process: detecting entities within individual maps and then associating them via a heuristic-based post-processing step. In this paper, we propose a novel approach that combines segmentation and association of geographic entities in historical maps using video instance segmentation (VIS). This method significantly streamlines geographic entity alignment and enhances automation. However, acquiring high-quality, video-format training data for VIS models is prohibitively expensive, especially for historical maps that often contain hundreds or thousands of geographic entities. To mitigate this challenge, we explore self-supervised learning (SSL) techniques to enhance VIS performance on historical maps. We evaluate the performance of VIS models under different pretraining configurations and introduce a novel method for generating synthetic videos from unlabeled historical map images for pretraining. Our proposed self-supervised VIS method substantially reduces the need for manual annotation. Experimental results demonstrate the superiority of the proposed self-supervised VIS approach, achieving a 24.9\% improvement in AP and a 0.23 increase in F1 score compared to the model trained from scratch. 
\end{abstract}

\section{Introduction}

Historical maps serve as unique and invaluable archives documenting the evolution of geographic features across diverse time periods \cite{Xia2023}. Numerous historical research endeavors necessitate synthesizing information derived from multiple historical maps, such as identifying the construction year of buildings, analyzing the development of road networks and settlements, supporting ecological and hydrological preservation and restoration efforts, and exploring the evolution of toponyms \cite{Heitzler2019}. Identifying and linking corresponding geographic entities from multiple historical maps enhances the structure and accessibility of historical map data \cite{Chiang2015}. In the cartographic domain, this issue is referred to as geographic entity alignment: when presented with a collection of historical maps $\{ \text{map}_1, \text{map}_2, \ldots, \text{map}_T \}$ illustrating the same geospatial location across various years, the goal is to identify matching entities across these maps that correspond to the same real-world features \cite{Sun2021a}. 

\begin{figure}
  \centering
  \includegraphics[width=1\linewidth]{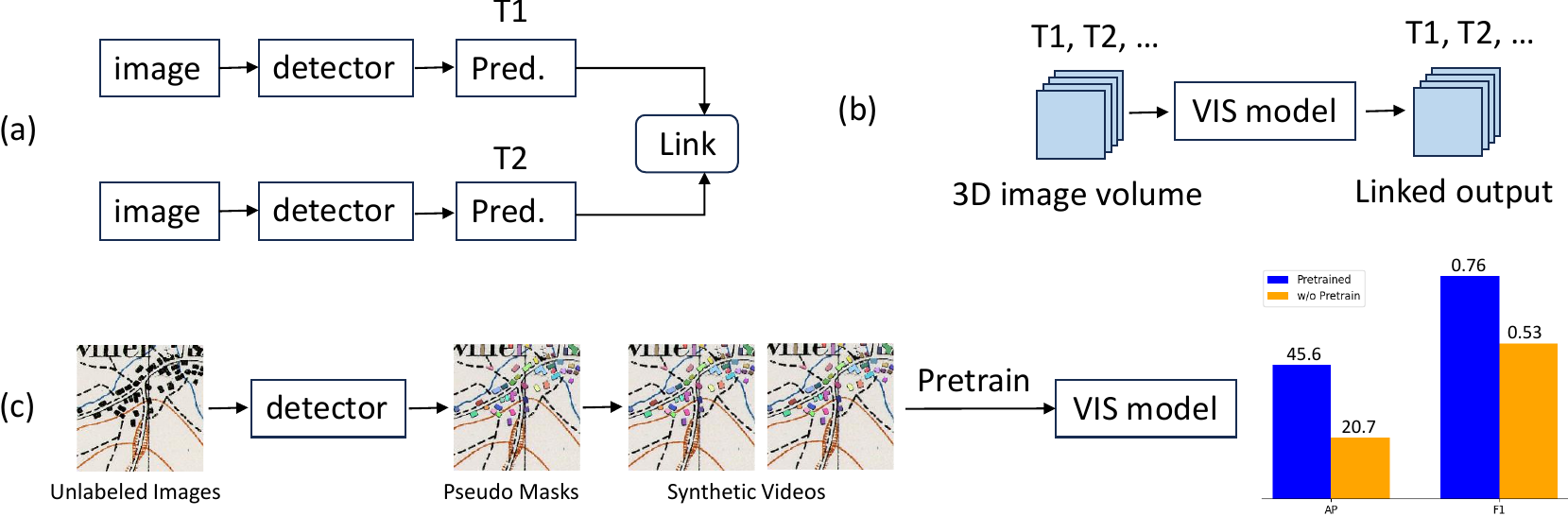}
  \caption{(a) Conventional approach for geographic entity alignment; (b) Proposed VIS method; (c) Proposed pretraining strategy for the VIS model.}
  \label{fig1}
\end{figure}

The conventional approach to geographic entity alignment involves two key steps (Figure \ref{fig1}(a)): 1. Extracting vector entities from scanned historical map images, typically achieved through image instance segmentation followed by vectorization; 2. Aligning corresponding entities from different maps using heuristic algorithms, including spatial distances, topological relations, and approximate topological relations \cite{Sun2021a}. The main limitation of the conventional approach is its low automation and reliance on handcrafted thresholds, compounded by varying distortions in historical maps across locations and time periods. 

To overcome these challenges, we propose a novel approach that integrates the segmentation and linking steps through the use of VIS, generating a 3D volume of linked entities directly (Figure \ref{fig1}(b)). The benefits of using the VIS method for linking geographic entities in historical maps are threefold: it enhances generalizability by eliminating the need for heuristic linking; it naturally tracks slight displacements caused by map distortions; and it significantly improves automation over conventional multistep methods.

Due to the scarcity and high cost of quality training data for VIS-based methods, we further explore the integration of SSL into the proposed pipeline. Many SSL methods for videos rely on motion-based learning signals, such as optical flow \cite{Yang2021, Xie2022}. However, these methods are not well-suited for stationary or slow-moving objects—common characteristics in historical maps, making them inadequate for this context. Although models are often initialized with weights pretrained on video segmentation benchmarks like YouTubeVIS-2019 \cite{Yang2019}, the significant domain gap between historical maps and natural videos limits their effectiveness when transferred to out-of-distribution tasks. To address this challenge, we propose a pretraining strategy by generating synthetic videos from unlabeled historical map images (Figure \ref{fig1}(c)). Experimental results demonstrate that the proposed pretraining strategy boosts VIS model performance on the downstream task by 24.9\% in AP and 0.23 in F1 score compared to training from scratch.

\section{Self-supervised VIS for Geographic Entity Alignment}
We present a self-supervised VIS approach for segmenting and linking geographic entities in historical maps. This method involves generating synthetic videos from unlabeled map images to pretrain the VIS model, which is then fine-tuned on downstream tasks. Further details are provided below.

\textbf{Generating Synthetic Videos from Maps.} The idea is to leverage large amounts of unlabeled historical map images to generate synthetic videos for pretraining. First, we use the small labeled dataset from the downstream task to train an image instance segmentation model, such as Mask R-CNN \cite{He2020}, to generate pseudo masks for the unlabeled images. We then create synthetic videos by duplicating the images and their pseudo masks to form two-frame sequences. Since the instance IDs are copied, objects remain naturally linked across the temporal dimension. Given that historical maps primarily feature static objects, this simple synthetic video approach effectively simulates real-world situations. Prior research shows that two-frame videos are sufficient to train a VIS model \cite{Wang2023}, so we limit our synthetic videos to two frames. This method enables us to generate nearly unlimited synthetic historical map videos for pretraining the VIS model without requiring additional manual annotations.

\textbf{Video Instance Segmentation.} Within the realm of VIS methods, two primary categories exist: the per-frame methods (a.k.a. online methods) and the per-clip methods (a.k.a. offline methods) \cite{Cheng2021}. Per-frame methods execute image segmentation independently on each frame and subsequently associate predicted instances across frames. In contrast, per-clip methods focus on modeling the spatio-temporal representations of individual video instances, enabling the direct prediction of a 3D mask for linked instances. In this paper, we adopt the per-clip VIS method, treating a historical map series $\{ \text{map}_1, \text{map}_2, \ldots, \text{map}_T \}$ as a 3D spatio-temporal volume of dimension $T \times H \times W$, where $T$ is the number of historical maps along the temporal axis, while $H$ and $W$ denote the height and width, respectively. To process the historical map sequence, we utilize Mask2Former-VIS \cite{Cheng2021}, which outputs a 3D volume containing linked building instances across time, i.e. $\{ e_1^i, e_2^i, \ldots, e_T^i \}$, where $e_t^i$ is the binary mask of a building entity $i$ in $\text{map}_t$. In Appendix \ref{appendix: Video Segmentation Model}, we further explain the architecture of the VIS model.

\textbf{Datasets and Evaluation Metrics.} This study utilizes two datasets: an unlabeled image dataset for pretraining and a labeled video dataset for fine-tuning. Both are generated from the Swiss Siegfried map (©swisstopo), with a focus on building instances. Image tiles from both datasets have a fixed dimension of 256 × 256 pixels. The unlabeled dataset contains 5,832 images, which are fed into Mask R-CNN to generate pseudo masks for building instances. Images without buildings in the pseudo masks are excluded from the subsequent process of generating synthetic videos, resulting in 4,384 synthetic videos for pretraining the VIS model. The labeled video dataset includes 497 training, 63 validation, and 62 testing videos, with each video consisting of four frames, corresponding to four timestamps (year 1896, 1904, 1932, and 1945) from the Siegfried map. The VIS dataset follows the YouTube-VIS \cite{Yang2019} format. We report Average Precision (AP) for videos, as used in YouTube-VIS \cite{Yang2019}, and F1 score as our evaluation metrics. Detailed definitions of these metrics are provided in Appendix \ref{appendix: Evaluation Metrics}.

\textbf{Implementation Details.} \textit{Supervised training of Mask R-CNN:} The model is initialized with pretrained weights from the COCO dataset \cite{Lin2015}. Training runs for 400 epochs with a batch size of 8 and a learning rate of 1e-3, conducted on a single Nvidia Quadro RTX 5000 GPU. \textit{Pretraining of Mask2Former-VIS:} We employ Mask2Former-VIS with a ResNet50 backbone, initializing the model using pretrained weights from the COCO dataset. The training process includes 4000 iterations with an initial learning rate of 1e-4, followed by a learning rate decay by a factor of 10 for an additional 2000 iterations.  This training is executed on 2 Nvidia Titan RTX GPUs with a batch size of 16. \textit{Fine-tuning on the downstream VIS task:} The Mask2Former-VIS model, pretrained on synthetic videos from the previous stage, is fine-tuned on the labeled video dataset. The fine-tuning follows the same configuration as the pretraining stage.

\section{Experimental Results}

\begin{table}[t!]
\centering
\caption{Mask2Former-VIS performance under different pretraining configurations}
\label{tab1}
\begin{tabular}{l | c c c c c c}
\toprule
\textbf{Pretraining Configurations} & \textbf{AP} & \textbf{AP50} & \textbf{AP75} & \textbf{Precision} & \textbf{Recall} & \textbf{F1} \\
\midrule
w/o Pretraining & 20.7 & 40.5 & 21.1 & 0.47 & 0.60 & 0.53 \\
COCO Images & \textbf{46.2} & \textbf{72.8} & \textbf{52.0} & \textbf{0.76} & \textbf{0.76} & \textbf{0.76} \\
YouTubeVIS-2019 & 44.8 & 69.8 & 49.8 & 0.74 & \textbf{0.76} & 0.75  \\
ImageNet Synthetic Videos & 32.0 & 56.5 & 33.5 & 0.64 & 0.68 & 0.66  \\
Map Synthetic Videos & 45.6 & 72.2 & 49.8 & 0.75 & \textbf{0.76} & \textbf{0.76} \\
% 0-shot Map Synthetic Videos & 17.9 & 39.1 & 15.9 & 0.41 & 0.61 & 0.49 \\
\bottomrule
\end{tabular}
\end{table}

To validate the effectiveness of the proposed self-supervised VIS method, we compare the performance of the Mask2Former-VIS model on the test set under different pretraining configurations. Table \ref{tab1} presents the numerical results. The results show that pretraining consistently improves model performance. Even pretraining with synthetic videos, following the VideoCutLER method \cite{Wang2023}, which uses only unlabeled ImageNet data to generate synthetic videos as pretraining samples, outperforms training from scratch. However, pretraining with labeled natural videos from the YouTubeVIS-2019 dataset yields better results than using synthetic videos from ImageNet. The best performance is achieved by models pretrained on COCO images and synthetic historical map videos.

The success of synthetic historical map videos over other video datasets is likely due to their greater semantic similarity to the downstream task data. However, the performance is on par with models pretrained on the COCO image instance segmentation dataset. This can be explained by the fact that Mask2Former-VIS \cite{Cheng2021} is an extension of an image instance segmentation model Mask2Former \cite{Cheng2022}. For single-frame input, Mask2Former operates as a standard image segmentation architecture, and for multi-frame video data, it simply shares queries across frames, enabling object segmentation and tracking. Due to this architectural design, Mask2Former-VIS does not benefit substantially from video-based pretraining. However, we emphasize that the proposed pretraining strategy could still prove advantageous for other VIS models that require video-based pretraining when applied to out-of-distribution historical map data.

\begin{figure}
  \centering
  \includegraphics[width=1\linewidth]{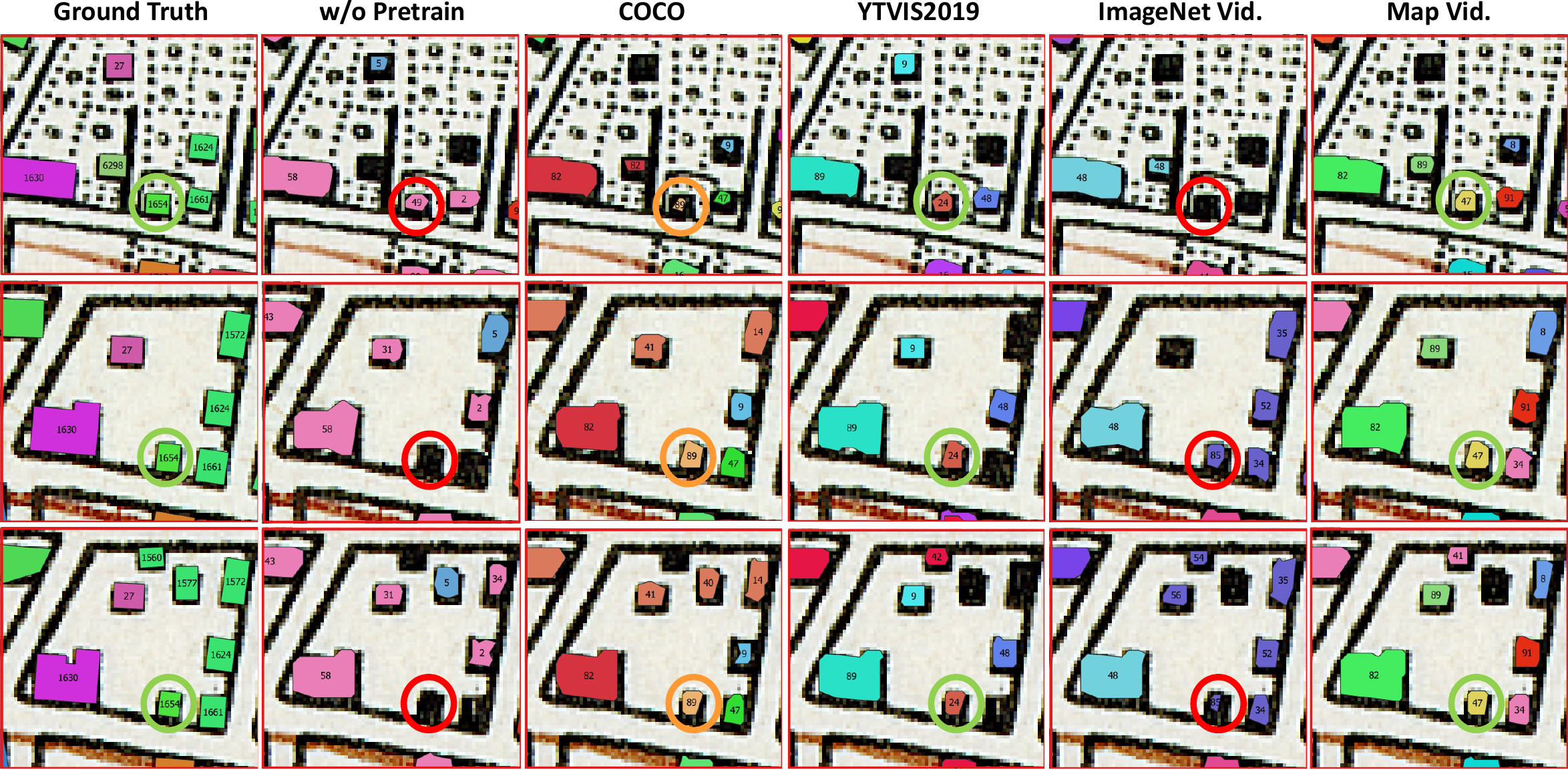}
  \caption{Segmentation and linking results for Mask2Former-VIS under different pretraining configurations. Results are shown for three frames corresponding to the years 1904, 1932, and 1945 (top to bottom).}
  \label{result}
\end{figure}

Figure \ref{result} visualizes the segmentation and linking results. Models without pretraining or those pretrained on ImageNet synthetic videos or the YouTubeVIS-2019 dataset show more missing detections in their visual output. In contrast, models pretrained on the COCO dataset and synthetic historical map videos demonstrate better segmentation performance, with the latter showing slightly improved geometric accuracy (e.g., the instance marked inside the circle). 

For instance tracking, all VIS models successfully track large instances or those without nearby confusing entities, regardless of pretraining configuration (e.g., the dark purple instance in the lower-left corner of the ground truth). However, in more complex scenarios—particularly when there are slight displacements of buildings (e.g., the light purple instance at the top of the ground truth)—the VIS models exhibit vulnerabilities, occasionally linking to the wrong entity. These scenarios can be challenging even for human annotators. Addressing this issue could be a key area for improvement, potentially by introducing slight displacements in the instances to simulate map distortions when generating synthetic historical map videos.

The numerical and visual results show that the proposed self-supervised VIS method offers an effective pipeline for automating geographic entity alignment. It addresses the inefficiencies associated with transferring pretrained weights to out-of-domain data, resulting in improved accuracy.

\section{Conclusion and Outlook}

This paper introduces a pioneering approach for segmenting and linking geographic entities in historical maps using self-supervised video instance segmentation. The proposed pretraining strategy generates large amounts of synthetic videos for use in the historical map domain, which differs significantly from natural videos and lacks sufficient labeled video data, without requiring additional manual annotations. It outperforms models pretrained on natural videos in linking building entities from Siegfried maps. Future work will focus on refining the strategy for generating synthetic historical map videos to better simulate real-world conditions, such as introducing slight displacements to mimic map distortions or merging smaller structures to reflect map generalization.

\begin{ack}
This research was funded by the Swiss National Science Foundation as part of the EMPHASES Project [Grant Number: 200021\_192018].
\end{ack}

\bibliographystyle{plain}
\bibliography{references.bib}

\appendix

\section{Video Segmentation Model}
\label{appendix: Video Segmentation Model}
\begin{figure}[h]
  \centering
  \includegraphics[width=0.8\linewidth]{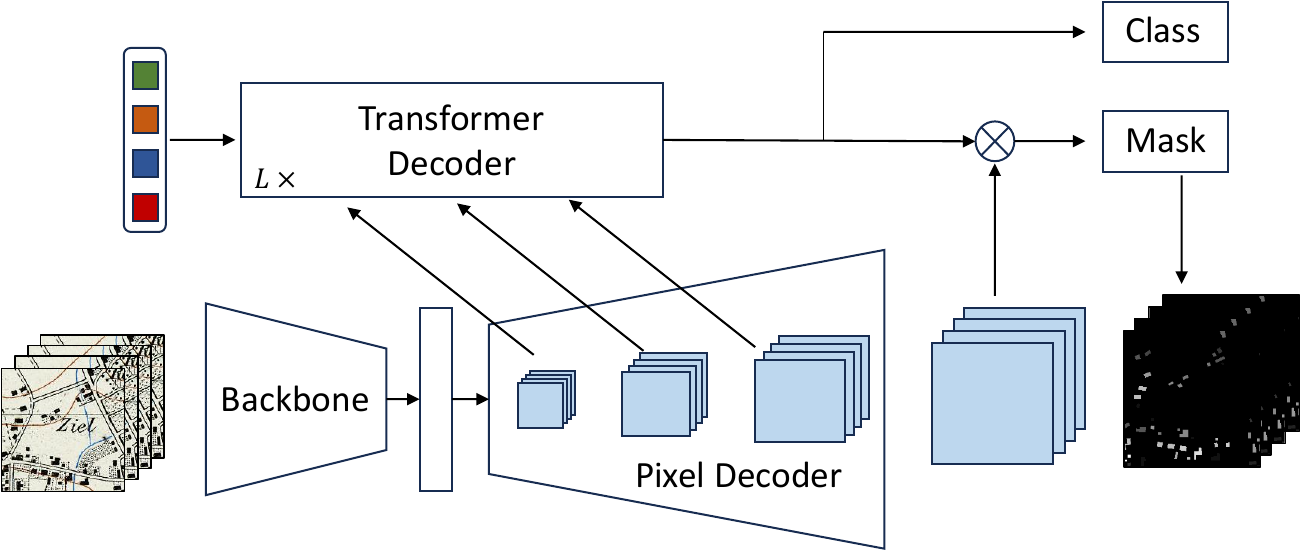}
  \caption{The simplified architecture of Mask2Former-VIS.}
  \label{model}
\end{figure}

The architecture of Mask2Former-VIS \cite{Cheng2021} is presented in Figure \ref{model}. It contains three major components: a backbone feature extractor, a pixel decoder, and a Transformer decoder. The backbone feature extractor extracts low-resolution features, which are later upsampled by the pixel decoder to generate multi-scale, high-resolution features. The Transformer decoder operates on the multi-scale features to process object queries. A linear classifier is then applied to the object queries, resulting in class probability predictions. The mask predictions are derived through a dot product between the query and the high-resolution features. Given that all features possess a 3D spatio-temporal volume and the queries are shared across frames along the temporal axis, the architecture effectively segments and tracks object instances across time.

\section{Evaluation Metrics}
\label{appendix: Evaluation Metrics}
Each linked instance $\{ e_1^i, e_2^i, e_3^i, e_4^i \}$ — representing four years and incorporating empty masks for non-existent instances — is considered a true positive (TP) if its Intersection over Union (IoU) with a corresponding ground truth instance exceeds 0.5. Predicted instances without a match are categorized as false positives (FP), while ground truth instances that go undetected are labeled as false negatives (FN). Unlike image instance segmentation, video instances consist of a sequence of masks, so IoU is calculated both spatially and temporally by summing the intersections and unions across all frames:
\begin{equation}
\text{IoU}(i, j) =
\frac{\sum_{t=1}^{4} \left| e_t^i \cap g_t^j \right|}
{\sum_{t=1}^{4} \left| e_t^i \cup g_t^j \right|}
\end{equation}
where $g_t^j$ stands for ground truth instance. After identifying TP, FP, and FN, precision, recall, and F1-score can be computed accordingly. AP is evaluated at 10 IoU thresholds, ranging from 50\% to 95\% with a step of 5\%.

\section{Traditional Approach}

\begin{table}[h]
\centering
\caption{Segmentation and linking results for Mask R-CNN}
\label{tab2}
\setlength{\extrarowheight}{1mm}
\begin{tabular}{l | c c c }
\toprule
\textbf{Image Instance Segmentation} & \textbf{AP} & \textbf{AP50} & \textbf{AP75} \\
 & 62.6 & 87.5 & 76.0 \\
\midrule
\textbf{Final Linking Results} & \textbf{P} & \textbf{R} & \textbf{F1} \\
 & 0.49 & 0.70 & 0.58 \\
\bottomrule
\end{tabular}
\end{table}

Conventional geographic entity alignment approaches first extract vector entities from single map sheets separately and then link the extracted entities between two maps. For a collection of three or more maps, the alignment is first performed between the maps of any two consecutive timestamps and then the results are combined \cite{Sun2021a}. Adhering to the two-step paradigm, we first use Mask R-CNN \cite{He2020} to effectively segment building instances from historical maps. Subsequently, we leverage approximate topological relations \cite{Clementini1997} to establish links between these instances. More precisely, we employ the "approximately within" relation, defining it as applicable when the intersection area between two entities exceeds or equals 60\% of the area of the smaller entity. The linked instances are then assigned a shared instance ID, ensuring uniqueness among different instances within a single image.

\begin{figure}[h]
  \centering
  \includegraphics[width=1\linewidth]{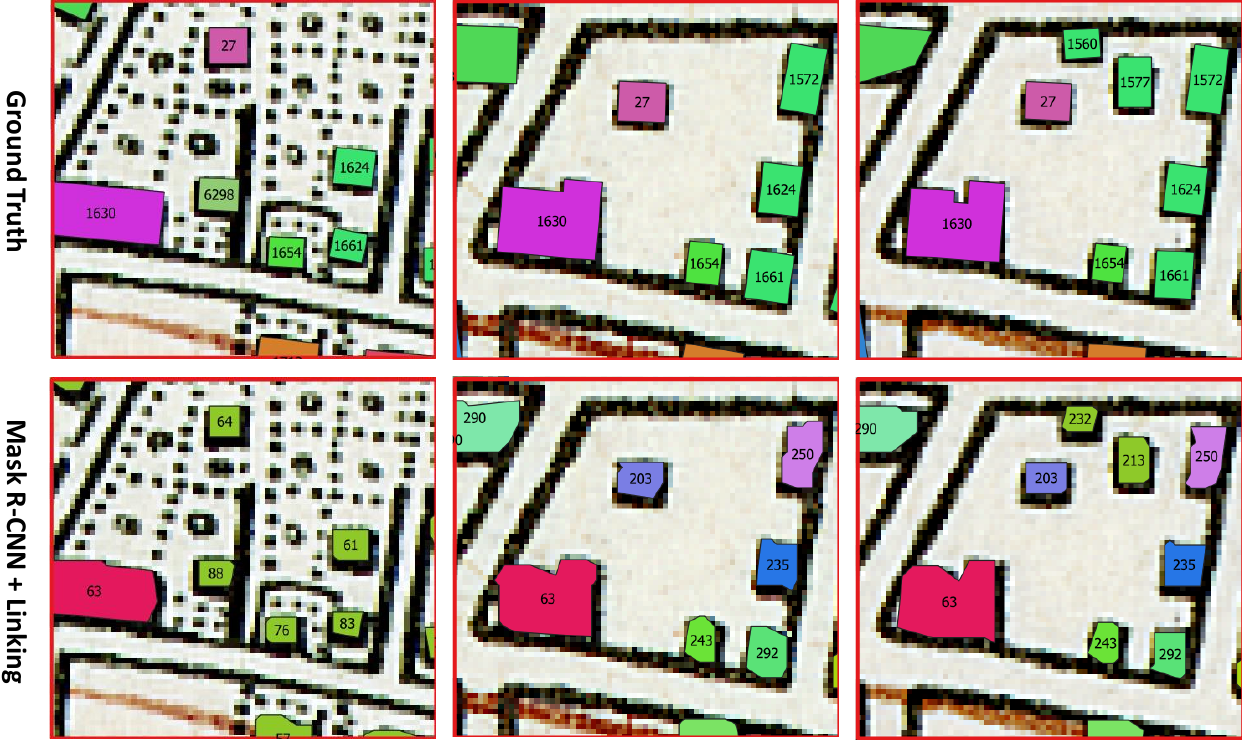}
  \caption{Segmentation and linking results for Mask R-CNN. Results are shown for three frames corresponding to the years 1904, 1932, and 1945 (left to right).}
  \label{maskrcnn}
\end{figure}

When exporting the Mask R-CNN output for building entity linking, the final result is fixed, and the confidence score is omitted. As a result, calculating the Average Precision (AP) for the final linked output of Mask R-CNN is not feasible. Therefore, we report only precision, recall, and F-score for the linked instances in Table \ref{tab2}. The results are visualized in Figure \ref{maskrcnn}.

Based on the numerical results, the proposed self-supervised VIS method outperformed the conventional two-step approach, achieving a 0.18 higher F1-score. The diminished performance of the conventional method is primarily due to its heuristic-based linking procedure, as evidenced in the visual outputs. While it performs well in associating entities in neighboring frames (e.g., year 1932 and 1945), it often fails to link corresponding entities over long sequences.

\end{document}